\documentclass{article}

\usepackage{arxiv}

\usepackage[utf8]{inputenc} 
\usepackage[T1]{fontenc}    
\usepackage{hyperref}       
\usepackage{url}            
\usepackage{tabularx}
\usepackage{pdflscape}
\usepackage{booktabs}       
\usepackage{amsfonts}       
\usepackage{amsmath}        
\usepackage{tikz}
\usetikzlibrary{positioning,decorations.pathreplacing}
\usepackage{nicefrac}       
\usepackage{microtype}      
\usepackage{lipsum}		
\usepackage{graphicx}
\usepackage{natbib}
\usepackage{doi}
\usepackage{pdflscape}
\usepackage{booktabs}
\usepackage{array}
\usepackage{ltablex}

\keepXColumns

\title{Mathematics for and by human cognition: \\A resource-rational search for bottlenecks in problem-solving}

\author{ \href{https://orcid.org/0000-0000-0000-0000}{\includegraphics[scale=0.06]{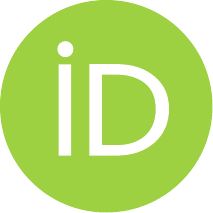}\hspace{1mm}Sneha Aenugu}\\
	Salk Institute of Biological Studies\\
	La Jolla, CA\\
	\texttt{saenugu@salk.edu} \\
}

\renewcommand{\shorttitle}{\textit{arXiv} Template}

\hypersetup{
pdftitle={A template for the arxiv style},
pdfsubject={q-bio.NC, q-bio.QM},
pdfauthor={David S.~Hippocampus, Elias D.~Striatum},
pdfkeywords={First keyword, Second keyword, More},
}

\begin{document}
\maketitle

\begin{abstract}

Human cognitive constraints are generally viewed as limiting factors in problem-solving. We argue that these constraints can instead play a critical role in driving advances in mathematics and beyond. We propose a theory of mathematical abstraction as a resource-rational search for bottlenecks in problem-solving. Bottlenecks arising from cognitive constraints create pressure to restructure existing knowledge, potentially giving rise to novel formalisms with applications beyond the problems that originally motivated them. Drawing on episodes from the history of mathematics, we illustrate how such bottlenecks can drive the development of novel abstractions and examine how cognitive constraints and affective responses shape this process. Finally, we discuss the implications of this account for machine mathematical discovery and argue that incorporating human-like constraints may facilitate the discovery of useful mathematical abstractions.

\end{abstract}


\section{Introduction}

The success of AI in solving several open problems has led to a reckoning in the field of mathematics. The incredible capabilities of AI have been celebrated by many as the key to solving several other open problems bugging humans --- climate change and cancer among many. Others have expressed dismay at the prospect of human mathematicians losing their relevance when confronted by the superhuman abilities of AI to consolidate information. In this article, we put forth a contrasting view. We claim that mathematics for and by human cognition is just as relevant as ever, both in advancing human understanding of the natural world and as an activity with an end unto itself. The cognitive limitations of humans are the central feature of this view.

Mathematics, like every human endeavor, is shaped by human cognitive limitations, the clues to which are sprinkled across its history. In this article, we take a dive into the past to identify some of the key transitions in the field of mathematics that can be understood as methodological innovations. These events largely follow a similar pattern. It starts with an unsolved problem or a computational bottleneck that leads to an \emph{impasse}, which is broken by the introduction of a novel formalism that renders it tractable. In the episodes between such methodological transitions, humans largely advance the field by exploring novel compositions of the existing toolkit to test its scope and limitations until they run into bottlenecks.

The bottlenecks that mathematicians face in problem-solving can be of two kinds: the problem itself can be ill-posed, or the problem can be well-posed but human cognitive limitations can place a barrier to its resolution. In the latter case, mathematicians resolve the bottlenecks by generating novel formalisms that overcome cognitive constraints and render the problems human-friendly. Unresolved problems are left as open questions for the field to grapple with for decades. Through case studies, we trace how mathematicians can generate novel formalisms that bypass their cognitive limitations to resolve bottlenecks in problem-solving.

We argue that human cognitive limitations are not bugs, but rather features in this regard. Capacity constraints create deliberate pressure to create novel formalisms that favor compression, efficiency, and composability. Humans can thus expand the repertoire of formalisms that they can manipulate to probe the natural world.

We discuss the implications of employing AI agents that lack human-like cognitive constraints in advancing mathematics.
\section{A resource-rational account of mathematical abstraction}

Human cognition is resource limited. Critically, our working memory is capacity constrained, with pivotal studies showing our inability to hold more than a few concepts in mind at a time \cite{Miller1956TheInformation, Cowan2001TheCapacity}. Our long-term memory is fragile as well, with concepts failing to be retrieved if they are not used frequently \cite{Bjork1992ANT}. Our reward seeking is also prone to discounting, whereby we favor sooner rewards over later ones, limiting our ability to optimize for long-term goals \cite{Frederick2002TimeReview}. We are susceptible to boredom when faced with tedious tasks and frustration when our persistent efforts fail to succeed. While the above features are limiting, normatively, they can be seen as adaptive when considering cognition under finite metabolic resources and the limited speed of biological information processing. This is the key principle behind \emph{resource rationality}. Our limited cognition is not a bug but a feature that enables us to perform efficiently given these constraints \cite{Callaway2022RationalPlanning, Lieder2020Resource-rationalResources, Bhui2021Resource-rationalMaking}. 

Descartes' musings in \emph{Discourse on Method} \cite{Descartes1950DiscourseMethod}, for instance, illustrate how the efficiency with which concepts could be represented and manipulated factored into his integration of geometrical and algebraic abstractions.

\begin{quote}
    Perceiving that in order to understand these relations I should sometimes have to consider them one by one, and sometimes only to bear them in mind, or embrace them in the aggregate, I thought that, in order the better to consider them individually, I should view them as subsisting between straight lines, than which I could find no objects more simple, or capable of being more distinctly represented to my imagination and senses; and on the other hand that in order to retain them in the memory, or embrace an aggregate of many, I should express them by certain characters, the briefest possible. In this way I believed that I could borrow all that was best both in geometrical analysis and in algebra, and correct all the defects of the one by help of the other.
\end{quote}

In this article, we extend this framing and claim that cognitive constraints are critical for facilitating breakthroughs in mathematical abstraction resulting in the generation of novel formalisms that are compact, efficient, and generalizable.
\subsection{Limited working memory as a feature driving compression through hierarchical abstraction}

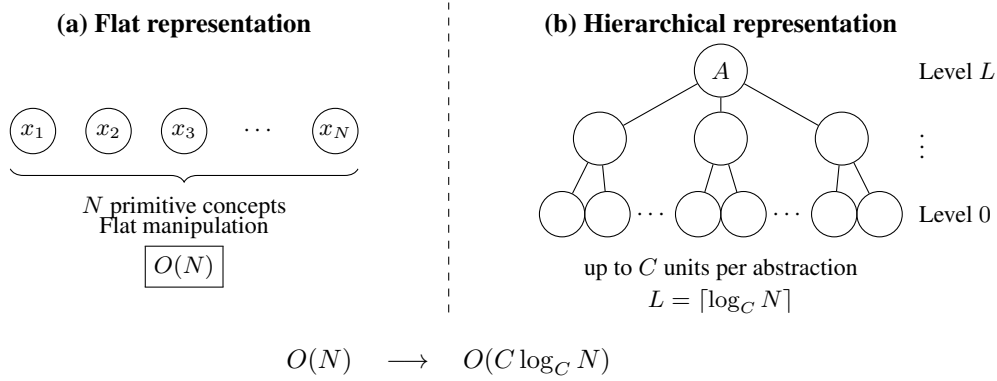
\begin{figure}[htpb!]
	\centering
	\begin{tikzpicture}[
		node distance=0.7cm,
		concept/.style={
			circle, draw, minimum size=6mm,
			inner sep=0pt, font=\small
		},
		abstract/.style={
			circle, draw, minimum size=7mm,
			inner sep=0pt
		},
		edge/.style={draw, thin},
		font=\small
	]
	
	
	\node[font=\bfseries] at (-3.5,3.2) {(a) Flat representation};
	
	\node[concept] (x1) at (-5.5,1.8) {$x_1$};
	\node[concept] (x2) at (-4.5,1.8) {$x_2$};
	\node[concept] (x3) at (-3.5,1.8) {$x_3$};
	
	\node at (-2.5,1.8) {$\cdots$};
	
	\node[concept] (xn) at (-1.5,1.8) {$x_N$};
	
	\draw[
		decorate,
		decoration={brace,mirror,amplitude=5pt}
	]
	(-5.8,1.3) -- (-1.2,1.3)
	node[midway,below=7pt] {$N$ primitive concepts};
	
	\node[align=center] at (-3.5,0.2)
	{Flat manipulation\\[2pt]
	$\boxed{O(N)}$};

	
	\draw[dashed] (0,-0.6) -- (0,3.5);

	
	\node[font=\bfseries] at (3.6,3.2)
	{(b) Hierarchical representation};
	
	\node[abstract] (root) at (3.6,2.6) {$A$};
	
	\node[abstract] (a1) at (2.0,1.7) {};
	\node[abstract] (a2) at (3.6,1.7) {};
	\node[abstract] (a3) at (5.2,1.7) {};
	
	\draw[edge] (root)--(a1);
	\draw[edge] (root)--(a2);
	\draw[edge] (root)--(a3);
	
	\node[concept] (b1) at (1.5,0.7) {};
	\node[concept] (b2) at (2.1,0.7) {};
	\node at (2.7,0.7) {$\cdots$};
	
	\node[concept] (b3) at (3.3,0.7) {};
	\node[concept] (b4) at (3.9,0.7) {};
	\node at (4.5,0.7) {$\cdots$};
	
	\node[concept] (b5) at (5.1,0.7) {};
	\node[concept] (b6) at (5.7,0.7) {};
	
	\draw[edge] (a1)--(b1);
	\draw[edge] (a1)--(b2);
	
	\draw[edge] (a2)--(b3);
	\draw[edge] (a2)--(b4);
	
	\draw[edge] (a3)--(b5);
	\draw[edge] (a3)--(b6);
	
	\node[right] at (6.1,2.6) {Level $L$};
	\node[right] at (6.1,1.7) {$\vdots$};
	\node[right] at (6.1,0.7) {Level $0$};
	
	\node[align=center] at (3.6,-0.25)
	{up to $C$ units per abstraction\\[2pt]
	$L = \lceil \log_C N\rceil$};
	
	\node[font=\bfseries] at (0,-1.25)
	{$O(N)\quad\longrightarrow\quad O(C \log_C N)$};
	
	\end{tikzpicture}
	
	\caption{
	Search among concepts with a flat representation requires direct manipulation of $N$ primitive
	concepts, whereas recursive abstraction packages up to $C$ concepts
	into reusable units, resulting a search complexity of $O(C \log_C N)$.
	}
	\label{fig:hierarchical-abstraction}
	\end{figure}

Through hierarchical abstraction, several low-level concepts can be compressed into a single higher-level concept, allowing a large problem space to be manipulated through a small number of abstract units. Consider a problem where one needs to search among $N$ primitive concepts. In a flat representation, if each concept must be individually considered, the search complexity scales linearly with the number of concepts, $O(N)$. Now suppose working memory imposes a capacity constraint $C$, such that only $C$ concepts can be searched among at a time. The search among $N$ concepts can be organized by hierarchically grouping them into $d$ levels, where $d \geq \log_C N$ (see Figure ~\ref{fig:hierarchical-abstraction}). To search among $N$ concepts grouped in this way, one can search across $d$ levels of the hierarchy and, at the final stage, search among at most $C$ concepts, yielding a search complexity of $O(C\log_C N)$. When $C$ is finite and small, the computational complexity can be greatly reduced, with $O(C\log_C N) \subseteq O(N)$.

The above examples show that deliberately imposing a capacity constraint and encouraging hierarchical abstraction can greatly reduce computational complexity. Thus, our natural cognitive limits can create a pressure that pushes us toward structuring concepts into computationally efficient hierarchies. This is just as well, as our brain does not run on billions of tokens or have access to vast computational resources; rather, it needs to perform its computations within a limited power budget of around 20 W. Moreover, regardless of energy efficiency, the ability to compress information into meaningful abstractions has itself been widely considered a marker of intelligence \cite{Huang2024CompressionLinearly, Chekaf2018CompressionIntelligence, Wolff2019InformationCognition, Schmidhuber2008DrivenJokes}.

\subsection{Emotions as a feature driving generation of novel abstractions}
The limited budget we have is to sustain multiple goals we pursue. We cannot spend numerous hours toiling away at a specific problem indiscriminately. Our emotions of tedium and frustration can come in handy here. Doing tedious computations can lead to fatigue or boredom that can pressure us into finding novel abstractions that can efficiently compress the problem space. An excerpt from Napier's \emph{Mirifici logarithmorum canonis descriptio} \cite{Napier1614MirificiDescriptio} shows the emotions that led to the development of \emph{logarithms} for efficient computations.
\begin{quote}
    …nothing is more tedious, fellow mathematicians, in the practice of the mathematical arts, than the great delays suffered in the tedium of lengthy multiplications and divisions, the finding of ratios, and in the extraction of square and cube roots… [with] the many slippery errors that can arise…I have found an amazing way of shortening the proceedings [in which]… all the numbers associated with the multiplications, and divisions of numbers, and with the long arduous tasks of extracting square and cube roots are themselves rejected from the work, and in their place other numbers are substituted, which perform the tasks of these rejected by means of addition, subtraction, and division ...
\end{quote}

Frustration is another emotion that can be adaptive in preventing us from venturing deeper into search spaces where complexity abounds. When a particular approach does not yield results despite repeated attempts with complex manipulations, we might be better off abandoning the approach in favor of constructing novel abstractions that can lead to simpler solutions. An excerpt from George Temple's speech titled \emph{The growth of mathematics} \cite{Temple1957Growth} clearly articulates this feature.

\begin{quote}
	The pattern of mathematical progress frequently exhibits the two stages of frustration and invention—frustration at the discovery that certain problems are insoluble with existing concepts and methods, and then the invention of new and more general concepts, new and more powerful methods to resolve these problems.
\end{quote}

Our emotions can thus be our assets and not liabilities by creating a pressure for generation of novel abstraction that avoid inefficient computations and searches while problem-solving.

\subsection{Resource-limited cognition favors reusable abstractions}

Memory retrieval is prone to failure, and for reliable memory access, concepts need to be practiced frequently \cite{Karpicke2007RepeatedRetention}. Building schemas that can be efficiently reused through analogical transfer across multiple domains can greatly reduce the cognitive load involved in encoding and retrieving individual concepts \cite{Gick1983SchemaTransfer}. The abstractions that tend to thrive in mathematical history are likely those that can be adopted with minimal transfer cost across several domains.

Hierarchical goal pursuit in problem solving helps stitch together disparate concepts to advance progress towards a specific goal \cite{Botvinick2008HierarchicallyPerspective}. The framework of \emph{lemmas} and \emph{theorems} can be seen as composing subproblems in service of a larger problem, where a given lemma or theorem can advance proofs for multiple problems. From a resource-rational perspective, mathematical abstractions with broad compositional utility are particularly valuable because a finite repertoire of learned abstractions can support an expanding space of problems. Abstractions like calculus are used so widely across multiple domains that they have become foundational to mathematical education. Most domains, in fact, aim to codify \emph{universal principles} that can explain a whole host of phenomena using a finite set of abstractions. Human resource rationality thus shapes mathematical abstraction towards the discovery of such \emph{universal principles}.

In the next section, we highlight the key forms of breakthroughs in mathematical abstraction that can be driven by resource-rational considerations.

\section{Resource-rational shaping of mathematical abstraction}

\subsection{Compression in symbolism} 

Compression is central to mathematical formalisms. Humans have long adopted symbols and notations that reduce complexity in representation. This can be traced back to the development of operators like multiplication, where repeated operations like

\begin{equation*}
     2 + 2 + 2 + 2 + 2 + 2 + 2
\end{equation*}

can be replaced by a compressed version:
\begin{equation*}
    2 \times 7,
\end{equation*}

where the operator $\times$ stands for the rule: repeat \emph{addition of 2} seven times.

Similar compression in notation can be seen in the development of matrix operations.

A system of $m$ linear equations in $n$ unknowns,
\[
\begin{aligned}
a_{11}x_1+\cdots+a_{1n}x_n &= b_1,\\
a_{21}x_1+\cdots+a_{2n}x_n &= b_2,\\
&\vdots\\
a_{m1}x_1+\cdots+a_{mn}x_n &= b_m,
\end{aligned}
\]
can be compressed into the matrix equation
\[
A\mathbf{x}=\mathbf{b},
\]
where
\[
A=(a_{ij})\in\mathbb{R}^{m\times n},\qquad
\mathbf{x}=(x_1,\ldots,x_n)^\top,\qquad
\mathbf{b}=(b_1,\ldots,b_m)^\top.
\]
Matrix multiplication packages the repeated operation of taking linear
combinations into a single algebraic operation:
\[
(A\mathbf{x})_i=\sum_{j=1}^{n}a_{ij}x_j.
\]
Thus, rather than manipulating $m$ equations and their coefficients
individually, the entire system can be treated as a single relation
between three mathematical objects, $A$, $\mathbf{x}$, and $\mathbf{b}$.
This representational compression enables higher-level operations on
the system as a whole, such as inversion, factorization, and
transformations of the coefficient matrix.

Often, individual symbolic elements are compressed together into higher-level abstractions, helping reduce cognitive overload and simplifying the complexity of describing and computing problems. 

\subsection{Lifting representations to new spaces}

Transforming the representational space in which a problem is framed is another form of abstraction that is often encountered in mathematical breakthroughs. One great example of this form is the Fourier representation. Originally formulated by Joseph Fourier while working on the heat equation, a partial differential equation for which solutions in general settings were difficult to obtain, the approach exploits the fact that specific cases like sinusoids can be solved more simply. This led to the representational change of expressing a given periodic function as a linear superposition of sinusoids. With this new abstraction, the heat equation could be solved for a much broader class of periodic functions. This particular innovation in representation generalized widely to a variety of problems whose formulations involved differential equations. 

Another instance of a representational change is the eigenvector formulation. Eigenvectors provide a representational innovation for simplifying linear transformations by identifying directions that remain invariant under the transformation. In an arbitrary coordinate system, a matrix $A$ can mix multiple components of a vector, making transformations such as $Ax$ and repeated applications $A^k x$ difficult to reason about. Eigenvectors identify special directions $v_i$ satisfying
\[
    Av_i = \lambda_i v_i,
\]
along which the transformation reduces to simple scalar multiplication. When $A$ has a complete basis of eigenvectors, this allows a change of coordinates
\[
    A = PDP^{-1},
    \qquad
    D = \operatorname{diag}(\lambda_1,\ldots,\lambda_n).
\]
In this representation, a coupled multidimensional transformation is decomposed into independent scalar transformations. For example,
\[
    A^k = PD^kP^{-1},
    \qquad
    D^k = \operatorname{diag}(\lambda_1^k,\ldots,\lambda_n^k).
\]
Thus, the eigenvector abstraction replaces reasoning about interactions among many coordinates with reasoning about independent modes of transformation, exposing invariant structure while simplifying repeated transformations and linear dynamical systems.

\subsection{Relational abstraction}

Mathematical objects have relations between them, and the structure of these relationships can be exploited to generate novel abstractions. A clear instance of this is the formalism of group theory. The origins of \emph{groups} can be traced in part to the problem of finding a solution to the general quintic. While there are precise formulas for calculating roots of quadratic, cubic, and quartic polynomials, a corresponding solution for the general quintic eluded mathematicians. This led to the development of the formalism of \emph{permutation groups}, whose elements describe permutations of the roots of a polynomial and allow one to study which permutations preserve algebraic relations among the roots and how these structures relate to the ability to express solutions in radicals. This concept of relational abstraction in the form of groups has been transferred to other formulations, such as \emph{Lie groups} and \emph{symmetry groups}, and the broader emphasis on relations and structure is foundational to fields such as \emph{category theory}.

\section{A resource-rational account of representational change}

The preceding examples suggest a common computational structure. A problem solver can either continue searching for a solution using their existing repertoire of abstractions or allocate resources toward changing the representation of the problem itself. This idea builds on theories of insight problem solving in which impasses can trigger representational change, allowing a solution that was inaccessible under the initial representation to become accessible \cite{Knoblich1999ConstraintSolving}. We extend this account by asking when such a shift becomes computationally worthwhile: we propose that bottlenecks provide a resource-rational signal for reallocating computation from search within the current representation toward search for a new one.

Let $\mathcal{R}$ denote the agent's current repertoire of abstractions, and
let $\pi=(r_1,\ldots,r_k)$, $r_i\in\mathcal{R}$, denote a candidate solution
constructed by composing elements of that repertoire. For a problem $X$, let
$C(\pi,X)$ denote the computational cost of finding and executing $\pi$.
This cost can reflect quantities such as the number of operations, search
effort, memory demand, or planning depth.

The minimum cost of solving $X$ under representation $\mathcal{R}$ is then

\[
C^\star(X;\mathcal{R})
=
\min_{\pi:\,\pi(X)\in G_X} C(\pi,X),
\]

where $G_X$ denotes the set of successful solutions. Importantly, a problem
may therefore be solvable in principle under $\mathcal{R}$ but computationally
inaccessible to a resource-limited agent.

\subsection{Bottlenecks reveal the cost of the current representation}

During problem solving, the agent does not know
$C^\star(X;\mathcal{R})$ in advance. Instead, it obtains evidence about the
cost of the current representation through the trajectory of its search.

Suppose computational step $t$ incurs cost $c_t$ and produces progress
$\Delta V_t$ toward the goal. The local return on computation is

\[
\eta_t=\frac{\Delta V_t}{c_t}.
\]

A bottleneck corresponds to a sustained decline in this return: additional
computation consumes resources without commensurate progress.

The phenomena discussed above can be understood as different manifestations
of this condition. \emph{Tedium} arises when long sequences of repetitive
operations accumulate execution costs for little additional progress.
\emph{Frustration} arises when repeated unsuccessful attempts accumulate
search costs without producing the expected progress. \emph{Planning failure}
occurs when no solution is found within the available planning horizon,
indicating that further search requires committing additional computational
resources.

These signals need not independently determine behavior. Rather, they provide
different sources of evidence for a common latent quantity: the expected cost
of continuing to solve the problem under the current representation.

\subsection{When should the representation change?}

The resource-rational decision is therefore a comparison between two uses of
computation. The agent can continue searching under $\mathcal{R}$, incurring
an expected remaining cost

\[
\mathbb{E}[C_{\mathrm{solve}}\mid X,\mathcal{R}],
\]

or allocate resources toward discovering a new abstraction $r$, producing an
expanded repertoire

\[
\mathcal{R}'=\mathcal{R}\cup\{r\}.
\]

Representational change is worthwhile when its expected discovery cost is
offset by the reduction in subsequent problem-solving cost:

\[
\mathbb{E}[C_{\mathrm{discover}}(r)
          + C^\star(X;\mathcal{R}\cup\{r\})]
<
\mathbb{E}[C_{\mathrm{solve}}\mid X,\mathcal{R}].
\]

This inequality captures the central proposal. Bottlenecks do not themselves
generate abstractions. Instead, they provide evidence that continued
computation under the current representation has low expected value, thereby
increasing the relative value of searching for a new representation.

A successful abstraction changes the computational structure of the problem.
For the problem that motivated its discovery, this means

\[
C^\star(X;\mathcal{R}\cup\{r\})
<
C^\star(X;\mathcal{R}).
\]

Its broader value depends on reuse. If the same abstraction reduces
problem-solving costs across a distribution of problems $X\sim p(X)$, then

\[
\mathbb{E}_{X\sim p(X)}
\left[
C^\star(X;\mathcal{R})
-
C^\star(X;\mathcal{R}\cup\{r\})
\right] > 0.
\]

Thus, an abstraction can repay the cost of its discovery not only by making
the immediate problem tractable, but also by reducing the cost of future problems
that share the same structure.

The proposed account can therefore be summarized as

\[
\text{costly search}
\;\longrightarrow\;
\text{bottleneck}
\;\longrightarrow\;
\text{representational search}
\;\longrightarrow\;
\text{lower-cost computation}.
\]

Our proposal is deliberately agnostic about the mechanism by which candidate
abstractions are generated. The claim is instead about the meta-level
allocation of computational resources: bottlenecks can make it
resource-rational to redirect computation away from searching for solutions
within an existing representation and toward searching for representations
in which solutions are cheaper to obtain.

In Appendix A~\ref{tab:case-studies}, we show a few instances from mathematical history in which key breakthroughs follow this pattern: the search for a solution to a problem within the existing mathematical repertoire imposes a bottleneck, which is then resolved by expanding the repertoire through the generation of a novel abstraction.






\section{Implications for machine mathematical discovery}

Large language models have been trained on centuries of mathematics shaped by human cognitive bottlenecks. As a result, they have inherited the representations, symbolism, and logic associated with this mathematics and can compose novel solutions to problems \cite{Novikov2025AlphaEvolve:Discovery}. However, the question of whether they can come up with novel abstractions beyond novel compositions of existing abstractions is still open. \cite{Zeng2026Nothing0} asked whether a large language model can come up with the abstraction of `zero' when pretrained on language. The authors concluded that a GPT-2 model pretrained on language could not create an abstraction of zero but could quickly learn it from a few examples. More research is advancing the ability of LLMs to form abstractions of problems using natural language. However, these attempts are limited to creating abstractions of the problem, with no reference to the novelty of the abstractions themselves \cite{Qu2025RLAD:Problems,ZhengTAKEMODELS}.

LLMs are known to generate programs with redundancy, often with code repeating the same functionality, and methods have been proposed to avoid this inefficiency through refactoring \cite{Stengel-Eskin2024ReGAL:Abstractions}. Studies investigating LLM reasoning show a propensity to overthink and excessively allocate computational resources toward solving simple questions \cite{Chen2024DoLLMs}. LLMs, unburdened by human cognitive bottlenecks, can indiscriminately allocate compute to problems by expanding the search space and potentially arriving at solutions. However, there is no free lunch, and someone has to pay the price. Earth's energy resources remain constrained, and relying on inefficient and indiscriminate searches might not scale to solving the problems of humanity.

Studies have attempted energy-aware computing in LLMs \cite{Han2024Token-Budget-AwareReasoning}; however, the solution might lie in human cognitive constraints. Evoking signals analogous to boredom and frustration in LLMs could prevent them from getting caught up in tedious computation and circuitous search spaces \cite{UrgenSchmidhuber1991AControllers}, prompting the meta-decision to expand the representational space instead. Nevertheless, the ability to discover novel abstractions might require architectural advancements beyond existing methods and perhaps cannot emerge through scale alone.

\section{Conclusion}

Human cognitive constraints and emotions are often seen as limitations to our reasoning abilities. Here, we put forth an alternative proposition: these human limitations can potentially create pressure for generating novel abstractions that drive mathematical breakthroughs. We propose that the process of mathematical abstraction can be viewed as a resource-rational search for bottlenecks in problem-solving. The meta-decision to switch from a strategy of composing solutions from an existing repertoire is modulated by the cognitive bottlenecks encountered during the process of problem-solving. We formalized three types of bottlenecks: tedium, frustration, and planning failure.

Studies in cognitive science have proposed the adaptive role of boredom as an information-seeking state that prompts the pursuit of novel experiences \cite{Bench2013OnBoredom, Bench2019BoredomExperiences, Geana2016UCExploration}. In mathematical problem-solving, several abstractions, such as logarithms, can be linked to the pressure to avoid tedious computations by restructuring representations. While the historical narratives in specific instances seem to support this hypothesis, a causal link between the emotion of boredom and the pressure to restructure is yet to be established. Other potential drivers of this pressure to compress could be related to opportunity costs in decision-making when choosing between multiple goals \cite{Otto2019TheEffort}.

Frustration, formalized as the difference between expected and realized progress, can also be adaptive in enabling disengagement from inefficient paths toward problem-solving \cite{Papini2022IncentiveNonreward}. Progress monitoring is thus a key heuristic in making the meta-decision of whether to persist with a given approach to a problem or switch to a different one \cite{MacGregor2001InformationProblems, Aenugu2025BuildingGoals, Aenugu2026WhyGoals}. While insufficient rates of progress can lead to the aversive state of feeling stuck, gamers and problem solvers have been shown to seek out these states \cite{Ross2024Impasse-DrivenStuck, MacGregor2001InformationProblems}. This also lends support to the hypothesis of open-ended exploration as a search for bottlenecks that need to be resolved.

Human working memory limits are crucial for generating hierarchical structures of abstraction \cite{Botvinick2008HierarchicallyPerspective}, where problems can be broken down into subproblems, and for discovering underlying latent structures in tasks \cite{Collins2013CognitiveStructure}. Working memory limits also play a role in how many concepts we can manipulate at a time and how far we can plan ahead \cite{Callaway2022RationalPlanning}, thereby reducing the accessible search space. This can prevent planning down arcane paths and guide search toward well-practiced and reusable routines. Upon planning failures, these limits can trigger a restructuring of the knowledge space instead of composing long-winded and inefficient solutions from available concepts.

We proposed a template for the discovery of novel abstractions, where bottleneck signals in problem-solving---frustration, tedium, and planning failure---can compound over time and trigger the meta-decision to quit executing a search over compositions of existing concepts and instead initiate a search over novel restructurings of knowledge spaces. In this article, we are conspicuously silent about the process by which novel abstractions are discovered, and that remains an active topic for future research. This direction, we believe, is key to creating energy-aware machine problem-solvers.

When machines solved one of the Millennium Prize Problems, Fields Medalists in mathematics signed a letter titled `A severe misalignment of AI in mathematics' \cite{DeclarationAI}. They emphasized the importance of the process of generating a solution rather than the solution itself. The theory of mathematical abstraction as a resource-rational search for bottlenecks emphasizes a similar point. Bottlenecks in problem-solving are critical junctures that force a restructuring of knowledge and generate novel abstractions that can expand the human mathematical repertoire. Struggling and failing are key motivators for triggering the restructuring of knowledge spaces. Devoid of these bottlenecks, the discovery of novel formalisms can be stalled, potentially holding back the generation of concepts in mathematics and beyond.

Human mathematicians remain extremely relevant now, just as they have been for centuries, in coming up with abstractions that are efficient and tractable. The idea that there is a single unique solution to a problem should be dispensed with, and a push should be made toward finding compressed solutions that are accessible. Any solution that is accessible to humans may tend to be computationally efficient and reusable. Efforts should be made to find solutions that can be understood by an inspired high school student. In this regard, mathematicians have their work cut out for them. Regardless of the push for compression, mathematics remains a siloed field, with only a few able to understand and appreciate its meaning. The barrier to entry into mathematics is high, given its lack of emphasis on public communication and the accessible dissemination of ideas. The recent foray of machines into the field of mathematics can perhaps serve as a reckoning with what is truly important.

\bibliographystyle{unsrtnat}
\bibliography{references-2}

\begin{landscape}

\section{Appendix}
\label{app:case-studies}

\small
\setlength{\tabcolsep}{3pt}
\renewcommand{\arraystretch}{1.1}
\setlength\LTleft{0pt}
\setlength\LTright{\fill}
\begin{tabularx}{0.92\linewidth}{
@{}
>{\raggedright\arraybackslash}p{2.2 cm}
*{4}{>{\raggedright\arraybackslash}X}
@{}
}

\caption{Mathematical innovations viewed through problems, bottlenecks, and outcomes.}
\label{tab:case-studies}\\

\toprule
\textbf{Case study} &
\textbf{Problem} &
\textbf{Bottleneck} &
\textbf{Expansion} &
\textbf{Outcome} \\
\midrule
\endfirsthead

\toprule
\textbf{Case study} &
\textbf{Problem} &
\textbf{Bottleneck} &
\textbf{Expansion} &
\textbf{Outcome} \\
\midrule
\endhead

\bottomrule
\endfoot

\textbf{Fourier series}
&
To solve the heat equation, a partial differential equation.
&
Solutions were known for simple functions like sinusoids, but complex equations remained intractable.
&
Express a complicated heat source as a superposition of simple sine and cosine waves.
&
Generalized to a wide array of mathematical and physical problems involving solving linear differential equations.
\\

\addlinespace[6pt]

\textbf{Complex numbers}
&
Solving cubic equations through the radical formula.
&
Solutions included roots of negative numbers, even though the equations have all real roots.
&
Introduce imaginary numbers $i$ as intermediate variables, which upon manipulation led to solutions in real numbers.
&
Generalized to several domains with different representational problems involving rotation, oscillations, and dynamical systems.
\\

\addlinespace[6pt]

\textbf{Cartesian (analytic) geometry}
&
Finding general methods for solving geometrical problems, for instance, locus problems such as the Pappus problem.
&
Each new problem required a new configuration of geometrical elements and their relations, making it difficult to generalize across problems. Moreover, complex problems posed a difficulty in representation.
&
Represent geometric quantities by algebraic variables and translate geometric relations into equations, allowing geometric problems to be manipulated using algebraic operations.
&
Unified algebra and geometry, provided general methods for representing and classifying curves, and made large classes of geometric problems accessible to systematic algebraic manipulation.
\\

\addlinespace[6pt]

\textbf{Group theory}
&
Finding solutions to polynomial equations of degree greater than 4.
&
After formulas were found for quadratics, cubics, and quartics, mathematicians repeatedly failed to find an analogous formula for the general quintic.
&
Replace the search for explicit roots with the study of transformations---permutations of the roots---that preserve algebraic relations among them. Encode these transformations and their composition as a group and relate the structural properties of this group to the possibility of solving the polynomial by radicals.
&
A problem of constructing solutions became a problem of classifying symmetry structures. The resulting abstraction generalized beyond polynomial equations into the modern theory of groups and became a fundamental language for symmetry across mathematics and physics.
\\

\addlinespace[6pt]

\textbf{Eigenvectors}
&
Understand a complicated linear transformation or repeated transformations $A^k x$.
&
Coordinates are coupled; applying $A$ mixes many variables simultaneously.
&
Search for invariant directions satisfying $Av=\lambda v$. Express the system in a basis of these natural directions. Matrix transformation becomes scalar multiplication $z_i\mapsto\lambda_i z_i$.
&
Dynamics, differential equations, stability, vibrations, Markov processes, PCA, etc., become easier to characterize.
\\

\addlinespace[6pt]


\textbf{Logarithms}
&
Perform multiplication, division, root extraction, and astronomical or trigonometric calculations with large numbers.
&
Long arithmetic calculations were laborious and error-prone.
&
Transform multiplication and division into addition and subtraction using logarithms and logarithmic tables.
&
Large numerical computations became substantially cheaper, and logarithms became a standard computational tool for centuries.
\\

\addlinespace[6pt]

\textbf{Zero / positional notation}
&
Represent arbitrary numbers and perform arithmetic efficiently.
&
Number systems without a positional placeholder require context-dependent symbols and cumbersome algorithms.
&
Treat zero as a placeholder within positional notation and eventually as a number participating in arithmetic.
&
Compact representation of arbitrarily large numbers and highly reusable written algorithms for arithmetic.
\\

\addlinespace[6pt]

\textbf{Non-Euclidean geometry}
&
Determine whether Euclid's parallel postulate could be derived from the remaining axioms.
&
Centuries of attempts to prove the fifth postulate from the others repeatedly failed.
&
Instead of continuing the proof search, alter the axiomatic representation by replacing the parallel postulate and explore the resulting geometries.
&
Hyperbolic and elliptic geometries demonstrated that alternative internally consistent geometric structures could be studied, showing that Euclid's parallel postulate is not a necessary consequence of the other Euclidean axioms.
\\

\addlinespace[6pt]

%
%

\end{tabularx}

\end{landscape}

\end{document}